# Non-iterative Coarse-to-fine Transformer Networks for Joint Affine and Deformable Image Registration

Mingyuan Meng[1,2], Lei Bi[2], Michael Fulham[1,3], Dagan Feng[1,4], and Jinman Kim[1]

[1] School of Computer Science, The University of Sydney, Sydney, Australia.
[2] Institute of Translational Medicine, Shanghai Jiao Tong University, Shanghai, China.
[3] Department of Molecular Imaging, Royal Prince Alfred Hospital, Sydney, Australia.
[4] Med-X Research Institute, Shanghai Jiao Tong University, Shanghai, China.
`lei.bi@sjtu.edu.cn`

**Abstract.** Image registration is a fundamental requirement for medical image analysis. Deep registration methods based on deep learning have been widely recognized for their capabilities to perform fast end-to-end registration. Many deep registration methods achieved state-of-the-art performance by performing coarse-to-fine registration, where multiple registration steps were iterated with cascaded networks. Recently, Non-Iterative Coarse-to-finE (NICE) registration methods have been proposed to perform coarse-to-fine registration in a single network and showed advantages in both registration accuracy and runtime. However, existing NICE registration methods mainly focus on deformable registration, while affine registration, a common prerequisite, is still reliant on time-consuming traditional optimization-based methods or extra affine registration networks. In addition, existing NICE registration methods are limited by the intrinsic locality of convolution operations. Transformers may address this limitation for their capabilities to capture long-range dependency, but the benefits of using transformers for NICE registration have not been explored. In this study, we propose a Non-Iterative Coarse-to-finE Transformer network (NICE-Trans) for image registration. Our NICE-Trans is the first deep registration method that (i) performs joint affine and deformable coarse-to-fine registration within a single network, and (ii) embeds transformers into a NICE registration framework to model long-range relevance between images. Extensive experiments with seven public datasets show that our NICE-Trans outperforms state-of-the-art registration methods on both registration accuracy and runtime.

## 1 Introduction

Image registration is a fundamental requirement for medical image analysis and has been an active research focus for decades [1]. It aims to find a spatial transformation between a pair of fixed and moving images, through which the moving image can be warped to spatially align with the fixed image. Similar to natural image registration [2], medical image registration usually requires affine registration to eliminate rigid misalignments and then performs additional deformable registration to address non-rigid deformations. Traditional methods usually formulate medical image registration as a time-consuming iterative optimization problem [3, 4]. Recently, deep registration methods based on deep learning have been widely adopted to perform end-to-end registration [5, 6]. Deep registration methods learn a mapping from image pairs to spatial transformations based on training data in an unsupervised manner, which have shown advantages in registration accuracy and computational efficiency [7-18].

Many deep registration methods perform coarse-to-fine registration to improve registration accuracy, where the registration is decoupled into multiple coarse-to-fine registration steps that are iteratively performed by using multiple cascaded networks [10-13] or repeatedly running a single network for multiple iterations [14, 15]. Mok et al. [13] proposed a Laplacian pyramid Image Registration Network (LapIRN), where multiple networks at different pyramid levels were cascaded. Shu et al. [14] proposed to use a single network (ULAE-net) to perform coarse-to-fine registration with multiple iterations. These methods perform iterative coarse-to-fine registration and extract image features repeatedly in each iteration, which inevitably increases computational loads and prolongs the registration runtime. Recently, Non-Iterative Coarse-to-finE (NICE) registration methods have been proposed to perform coarse-to-fine registration with a single network in a single iteration [16-18]. For example, we previously proposed a NICE registration network (NICE-Net) [18, 19], where multiple coarse-to-fine registration steps are performed with a single network in a single iteration. These NICE registration methods show advantages in both registration accuracy and runtime on the benchmark task of intra-patient brain MRI registration. Nevertheless, we identified that existing NICE registration methods still have two main limitations.

Firstly, existing NICE registration methods merely focus on deformable coarse-to-fine registration, while affine registration, a common prerequisite, is still reliant on traditional registration methods [16, 18] or extra affine registration networks [17]. Using traditional registration methods incurs time-consuming iterative optimization, while cascading extra networks consumes additional computational resources (e.g., extra GPU memory and runtime). Secondly, existing NICE registration methods are based on Convolution Neural Networks (CNN) and thus are limited by the intrinsic locality (i.e., limited receptive field) of convolution operations. Transformers have been widely adopted in many medical applications for their capabilities to capture long-range dependency [20]. Recently, transformers have also been shown to improve registration with conventional Voxelmorph [7]-like architecture [21-23]. However, the benefits of using transformers for NICE registration have not been explored.

In this study, we propose a Non-Iterative Coarse-to-finE Transformer network (NICE-Trans) for joint affine and deformable registration. Our technical contributions are two folds: (i) We extend the existing NICE registration framework to affine registration, where multiple steps of both affine and deformable coarse-to-fine registration are performed with a single network in a single iteration. (ii) We explore the benefits of transformers for NICE registration, where Swin Transformer [24] is embedded into the NICE-Trans to model long-range relevance between fixed and moving images. This is the first deep registration method that integrates previously separated affine and deformable coarse-to-fine registration into a single network, and this is also the first deep registration method that exploits transformers for NICE registration. Extensive experiments with seven public datasets show that our NICE-Trans outperforms state-of-the-art registration methods on both registration accuracy and runtime.

## 2   Method

Image registration aims to find a spatial transformation $\phi$ that warps a moving image $I_m$ to a fixed image $I_f$, so that the warped image $I_{m \circ \phi} = I_m \circ \phi$ is spatially aligned with the $I_f$. In this study, we assume the $I_m$ and $I_f$ are two single-channel, grayscale volumes defined in a 3D spatial domain $\Omega \subset \mathbb{R}^3$, which is consistent with common medical image registration studies [7-18]. The $\phi$ is parameterized as a displacement field, and we parametrized the image registration problem as a function $\mathcal{R}_\theta(I_f, I_m) = \phi$ using NICE-Trans. As shown in Fig. 1, our NICE-Trans consists of an intra-image feature learning encoder and an inter-image relevance modeling decoder (refer to Section 2.1). Multiple steps of affine and deformable registration are performed within a single network iteration (refer to Section 2.2). The $\theta$ is a set of learnable parameters that are optimized through unsupervised learning (refer to Section 2.3).

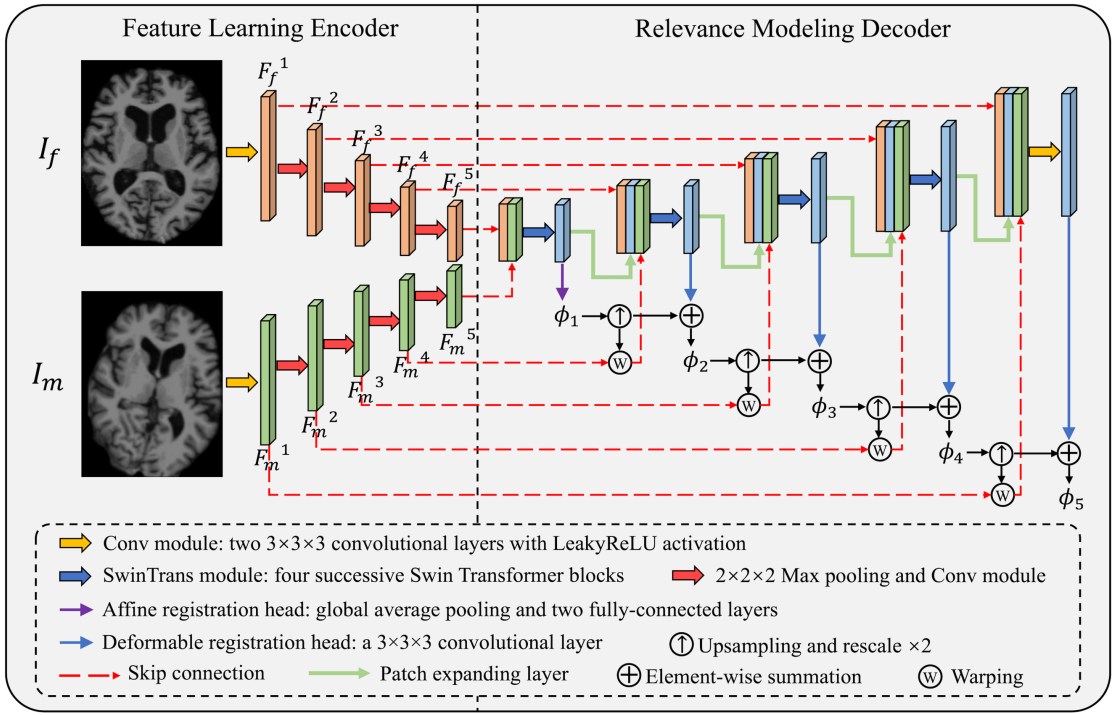

**Fig. 1.** The Architecture of our NICE-Trans. The affine and deformable registration steps, $L_a$ and $L_d$, are set as 1 and 4 for illustration.

### 2.1   Non-iterative Coarse-to-fine Transformer Networks (NICE-Trans)

The architecture of the proposed NICE-Trans is presented in Fig. 1, which consists of a dual-path encoder to learn image features from $I_m$ and $I_f$ separately and a single-path decoder to model the spatial relevance between $I_m$ and $I_f$. Skip connections are used at multiple scales to propagate features from the encoder to the decoder. Here, we assume the NICE-Trans performs $L_a$ and $L_d$ steps of affine and deformable registration, resulting in a total of $L = L_a + L_d$ steps of coarse-to-fine registration.

The encoder has two identical, weight-shared paths $P_m$ and $P_f$ that take $I_m$ and $I_f$ as input, respectively. Each path consists of $L$ successive Conv modules with 2×2×2 max pooling applied between two adjacent modules, which produces two $L$-level feature pyramids $F_m \in \{F_m^1, F_m^2, ..., F_m^L\}$ and $F_f \in \{F_f^1, F_f^2, ..., F_f^L\}$, where the $F_f^i$ and $F_m^i$ are the output of the $i^{th}$ Conv module in the $P_f$ and $P_m$. Each Conv module consists of two 3×3×3 convolutional layers followed by LeakyReLU activation with parameter 0.2. This dual-path design can learn uncoupled image features of $I_m$ and $I_f$, which enables the NICE-Trans to reuse the learned features at multiple registration steps, thereby discarding the requirement for repeated feature learning.

The decoder consists of $L$-1 SwinTrans modules and a Conv module, with a patch expanding layer [23] applied between two adjacent modules to double the feature resolution and halve the feature dimension. Each SwinTrans module consists of one 1×1×1 convolutional layer for feature dimension reduction and four successive Swin Transformer blocks [24] including layer normalization, Window/Shifted Window-based Multi-head Self-Attention (W/SW-MSA), Multilayer Perceptron (MLP), and residual connections. The output of each decoder module is fed into an affine or deformable registration head that maps the input features into a displacement field, which produces $L$ displacement fields $\phi_i \in \{\phi_1, \phi_2, ..., \phi_L\}$ for $L$ steps of coarse-to-fine registration (detailed in Section 2.2). The output of each patch expanding layer is concatenated with $F_f^i$ and $F_m^i \circ \phi_{i-1}$, which is then fed into its later decoder module. The decoder performs finer registration after each decoder module, where the $\phi_L$ is the final output $\phi$. Detailed architecture settings (e.g., feature dimensions, head numbers of self-attention) are presented in the supplementary materials.

Our NICE-Trans differs from the existing NICE-Net [18] mainly in two aspects: (i) our NICE-Trans integrates affine and deformable registration into a unified network, and (ii) our NICE-Trans leverages Swin Transformer to model long-range spatial relevance between $I_m$ and $I_f$. In addition, the existing NICE-Net extracts features from the intermediately warped image at each registration step, while our NICE-Trans directly warps the $F_m$ to avoid this process and achieves similar performance.

## 2.2 Joint Affine and Deformable Registration

The output features of the first $L_a$ decoder modules are fed into $L_a$ affine registration heads, where the features are mapped to a 3×4 affine matrix through global average pooling and two fully-connected layers, which are then sampled as a dense displacement field. After the first $L_a$ steps of affine registration, the output features of the last $L_d$ decoder modules are fed into $L_d$ deformable registration heads, where the features are directly mapped to a dense displacement field via a 3×3×3 convolutional layer.

At the beginning of coarse-to-fine registration, the $\phi_1$ is the output of the first registration head. Then, the $\phi_1$ is upsampled (×2) and voxel-wisely added to the output of the second registration head to derive $\phi_2$. This process is repeated until the $\phi_L$ is derived, which realizes joint affine and deformable coarse-to-fine registration. In our experiments, we set $L_a$ and $L_d$ as 1 and 4 (illustrated in Fig.1) as this setting achieved the best validation results (refer to the supplementary materials). Fig. 2 exemplifies a registration result of the NICE-Trans with five steps of coarse-to-fine registration.

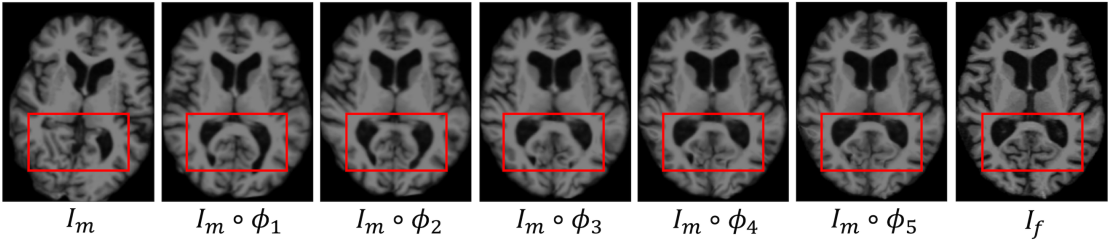

**Fig. 2.** Registration results of the NICE-Trans with $L_a = 1$ and $L_d = 4$. From left to right are the moving image, the images warped by 5 registration steps, and the fixed image.

### 2.3 Unsupervised Learning

The learnable parameters $\theta$ are optimized using an unsupervised loss $\mathcal{L}$ that does not require labels. The $\mathcal{L}$ is defined as $\mathcal{L} = \mathcal{L}_{sim} + \sigma \mathcal{L}_{reg}$, where the $\mathcal{L}_{sim}$ is an image similarity term that penalizes the differences between the warped image $I_{m \circ \phi}$ and the fixed image $I_f$, the $\mathcal{L}_{reg}$ is a regularization term that encourages smooth and invertible transformations $\phi$, and the $\sigma$ is a regularization parameter.

We adopt negative local normalized cross-correlation (NCC) as the $\mathcal{L}_{sim}$, which is a widely used similarity metric in image registration methods [7-10, 12-18]. For the $\mathcal{L}_{reg}$, we impose a diffusion regularizer on the $\phi$ to encourage its smoothness and also adopt a Jacobian Determinant (JD) loss [25] to enhance its invertibility. As the $\phi$ is not invertible at voxel $p$ where the Jacobian determinant is negative ($|J\phi(p)| \leq 0$) [26], the JD loss explicitly penalizes the negative Jacobian determinants of $\phi$. Finally, the $\mathcal{L}_{reg}$ is defined as $\mathcal{L}_{reg} = \sum_{p \in \Omega} ||\nabla \phi(\boldsymbol{p})||^2 + \lambda JD(\phi)$, where the $\lambda$ is a regularization parameter balancing registration accuracy and transformation invertibility.

## 3 Experimental Setup

### 3.1 Dataset and Preprocessing

We evaluated the proposed NICE-Trans on the task of inter-patient brain MRI registration, which is a common benchmark task in medical image registration studies [7-9, 12-18]. We followed the dataset settings in [18]: 2,656 brain MRI images acquired from four public datasets (ADNI [27], ABIDE [28], ADHD [29], and IXI [30]) were used for training; two public brain MRI datasets with anatomical segmentation (Mindboggle [31] and Buckner [32]) were used for validation and testing. The Mindboggle dataset contains 100 MRI images and were randomly split into 50/50 images for validation/testing. The Buckner dataset contains 40 MRI images and were used for testing only. In addition to the original settings of [18], we adopted an additional public brain MRI dataset (LPBA [33]) for testing, which contains 40 MRI images.

We performed brain extraction and intensity normalization for each MRI image with FreeSurfer [32]. Each image was placed at the same position via Center of Mass (CoM) initialization [34], and then was cropped into 144×192×160 voxels.

## 3.2 Implementation Details

We implemented our NICE-Trans using PyTorch on a NVIDIA Titan V GPU with 12 GB memory. We used an ADAM optimizer with a learning rate of 0.0001 and a batch size of 1 to train the NICE-Trans for 100,000 iterations. At each iteration, two images were randomly picked from the training data as the fixed and moving images. A total of 100 image pairs, randomly picked from the validation data, were used to monitor the training process and to optimize hyper-parameters. We set $\sigma$ as 1 to ensure that the $\mathcal{L}_{sim}$ and $\sigma\mathcal{L}_{reg}$ have close values, while the $\lambda$ was set as $10^{-4}$ to ensure that the percentage of voxels with negative Jacobian determinants is less than 0.05% (refer to the supplementary materials for detailed regularization analysis). Our code will be available in https://github.com/MungoMeng/Registration-NICE-Trans.

## 3.3 Comparison Methods

Our NICE-Trans was compared with nine image registration methods, including two traditional methods and seven deep registration methods. The compared traditional methods are SyN [3] and NiftyReg [4]. For these methods, we used cross-correlation as the similarity measure and adopted FLIRT [35] for affine registration. The compared deep registration methods are VoxelMorph (VM) [7], Diffeomorphic VoxelMorph (DifVM) [8], TransMorph [21], Swin-VoxelMorph (Swin-VM) [22], LapIRN [13], ULAE-net [14], and NICE-Net [18]. The VM and DifVM are two commonly benchmarked registration methods in the literature [12-18, 21-23]. The TransMorph and Swin-VM are two state-of-the-art methods that embed Swin Transformer into VM-like architecture. The LapIRN, ULAE-net, and NICE-Net are three state-of-the-art coarse-to-fine registration methods. For the compared deep registration methods, we adopted NCC as the similarity loss and followed [17, 36] to cascade a CNN-based registration network (AffineNet) for affine registration.

## 3.4 Experimental Settings

We compared the NICE-Net to the nine comparison methods for subject-to-subject registration. For testing, we randomly picked 100 image pairs from each of the Mindboggle, Buckner, and LPBA testing sets. We used standard evaluation metrics for medical image registration [7-18]. The registration accuracy was evaluated using the Dice similarity coefficients (DSC) of segmentation labels, while the smoothness and invertibility of spatial transformations were evaluated using the percentage of Negative Jacobian Determinants (NJD). Generally, a higher DSC and a lower NJD indicate better registration performance. A two-sided *P* value less than 0.05 is considered to indicate a statistically significant difference between two DSCs.

We also performed an ablation study to explore the benefits of transformers. We built a baseline method that has the same architecture as the NICE-Trans but only uses Conv modules. After that, we embedded Swin Transformer into the baseline method, where SwinTrans modules replaced the Conv modules in the encoder (Trans-Encoder), decoder (Trans-Decoder), or both (Trans-All).

**Table 1.** Registration performance of our NICE-Trans and all comparison methods.

| Method | | Mindboggle | | Buckner | | LPBA | | Runtime (s) | |
|---|---|---|---|---|---|---|---|---|---|
| | | DSC | NJD (%) | DSC | NJD (%) | DSC | NJD (%) | CPU | GPU |
| Before registration | | 0.269* | / | 0.330* | / | 0.536* | / | / | / |
| FLIRT (affine only) | | 0.347* | / | 0.406* | / | 0.626* | / | 58.2 | / |
| SyN (no affine) | | 0.535* | 0.25 | 0.566* | 0.28 | 0.674* | 0.12 | 3688 | / |
| NiftyReg (no affine) | | 0.558* | 0.32 | 0.601* | 0.35 | 0.690* | 0.15 | 165 | / |
| FLIRT + | SyN | 0.548* | 0.26 | 0.577* | 0.25 | 0.692* | 0.09 | 3746 | / |
| | NiftyReg | 0.567* | 0.34 | 0.610* | 0.30 | 0.705* | 0.13 | 223 | / |
| AffineNet (affine only) | | 0.341* | / | 0.400* | / | 0.611* | / | 1.12 | 0.118 |
| VM (no affine) | | 0.518* | 2.63 | 0.558* | 2.37 | 0.663* | 1.21 | 3.85 | 0.395 |
| DifVM (no affine) | | 0.502* | 0.042 | 0.548* | 0.032 | 0.671* | 0.005 | 3.92 | 0.446 |
| TransMorph (no affine) | | 0.545* | 2.25 | 0.585* | 2.15 | 0.682* | 1.27 | 3.90 | 0.432 |
| Swin-VM (no affine) | | 0.542* | 0.022 | 0.589* | 0.017 | 0.684* | 0.004 | 5.82 | 0.550 |
| LapIRN (no affine) | | 0.563* | 0.046 | 0.599* | 0.039 | 0.688* | 0.006 | 6.52 | 0.624 |
| ULAE-net (no affine) | | 0.579* | 2.08 | 0.611* | 2.00 | 0.695* | 1.06 | 7.21 | 0.730 |
| NICE-Net (no affine) | | 0.580* | 0.048 | 0.611* | 0.034 | 0.696* | 0.004 | 4.17 | 0.423 |
| AffineNet + | VM | 0.548* | 2.54 | 0.580* | 2.24 | 0.682* | 1.17 | 4.97 | 0.513 |
| | DifVM | 0.526* | 0.048 | 0.565* | 0.027 | 0.686* | 0.005 | 5.04 | 0.564 |
| | TransMorph | 0.568* | 2.14 | 0.604* | 2.18 | 0.694* | 1.10 | 5.02 | 0.550 |
| | Swin-VM | 0.563* | 0.024 | 0.607* | 0.021 | 0.696* | 0.003 | 6.94 | 0.668 |
| | LapIRN | 0.581* | 0.042 | 0.611* | 0.036 | 0.699* | 0.006 | 7.64 | 0.742 |
| | ULAE-net | 0.595* | 2.12 | 0.625* | 1.92 | 0.705* | 0.97 | 8.33 | 0.848 |
| | NICE-Net | 0.596* | 0.034 | 0.624* | 0.026 | 0.705* | 0.004 | 5.29 | 0.541 |
| NICE-Trans (affine only) | | 0.353* | / | 0.410* | / | 0.618* | / | 1.04 | 0.105 |
| NICE-Trans (no affine) | | 0.594* | 0.018 | 0.622* | 0.016 | 0.704* | 0.003 | 4.52 | 0.480 |
| NICE-Trans (ours) | | **0.612** | **0.016** | **0.636** | **0.015** | **0.715** | **0.002** | **4.69** | **0.486** |

**Bold**: the best DSC and NJD in each testing dataset and the shortest runtime of completing both affine and deformable registration. *: $P<0.05$, in comparison to NICE-Trans (ours).

## 4   Results and Discussion

Table 1 presents the registration performance of our NICE-Trans and all comparison methods. The registration accuracy of all methods degraded by 1-3% in DSC when affine registration was not performed, which demonstrates the importance of affine registration. However, using FLIRT or AffineNet for affine registration incurred extra computational loads and increased the registration runtime. Our NICE-Trans performed joint affine and deformable registration, which enabled it to realize affine registration with negligible additional runtime. Moreover, we suggest that integrating affine and deformable registration into a single network also brings convenience for network training. Training two separate affine and deformable registration networks will prolong the whole training time, while joint training will consume more GPU memory. As for registration accuracy, the TransMorph and Swin-VM achieved higher DSCs than the conventional VM and DifVM, but still cannot outperform the existing CNN-based coarse-to-fine registration methods (LapIRN, ULAE-net, and NICE-Net). Our NICE-Trans leverages Swin Transformer to perform coarse-to-fine registration, which enabled it to achieve the highest DSCs among all methods. This means that our NICE-Trans also has advantages on registration accuracy. We present a qualitative comparison in the supplementary materials, which shows that the registration result produced by our NICE-Trans is more consistent with the fixed image. In addition, there usually exists a trade-off between DSC and NJD as imposing constraints on the spatial transformations limits their flexibility, which results in degraded registration accuracy [13, 18]. For example, compared with VM, the DifVM with diffeomorphic constraints achieved better NJDs and worse DSCs. Nevertheless, our NICE-Trans achieved both the best DSCs and NJDs. We suggest that, if we set $\lambda$ as 0 to maximize the registration accuracy with the cost of transformation invertibility, our NICE-Trans can achieve higher DSCs and outperform the comparison methods by a larger margin (refer to the regularization analysis in the supplementary materials).

**Table 2.** Results of our ablation study.

| Method | Mindboggle | | Buckner | | LPBA | | Runtime (s) | |
|---|---|---|---|---|---|---|---|---|
| | DSC | NJD (%) | DSC | NJD (%) | DSC | NJD (%) | CPU | GPU |
| Baseline | 0.600 | 0.028 | 0.627 | 0.025 | 0.706 | 0.003 | 4.25 | 0.438 |
| Trans-Encoder | 0.600 | 0.024 | 0.625 | 0.018 | 0.705 | 0.003 | 4.82 | 0.488 |
| Trans-Decoder (ours) | **0.612** | 0.016 | **0.636** | 0.015 | **0.715** | 0.002 | 4.69 | 0.486 |
| Trans-All | **0.612** | **0.015** | 0.634 | 0.017 | 0.714 | **0.002** | 5.48 | 0.557 |

**Bold**: the best DSC and NJD in each testing dataset.

Table 2 shows the results of our ablation study. Swin Transformer improved the registration performance when embedded into the decoder, but had limited benefits in the encoder. This suggests that Swin Transformer can benefit registration in modeling inter-image spatial relevance while having limited benefits in learning intra-image representations. This finding is intuitive as image registration aims to find spatial relevance between images, instead of finding the internal relevance within an image. Under this aim, embedding transformers in the decoder helps to capture long-range relevance between images and improves registration performance. We noticed that previous studies gained improvements by embedding Swin Transformer in the encoder [21] or leveraging a full transformer network [22]. This is attributed to the fact that they used a VM-like architecture that entangles image representation learning and spatial relevance modeling throughout the whole network. Our NICE-Trans decouples these two parts and provides further insight on using transformers for registration: leveraging transformers to learn intra-image relevance might not be beneficial but merely incurs extra computational loads.

It should be acknowledged that there are a few limitations in our study. First, the experiment (Table 1) demonstrated that our NICE-Trans can well address the inherent misalignments among inter-patient brain MRI images, but the sensitivity of affine registration to different degrees of misalignments is still awaiting further exploration. Second, in this study, we evaluated the NICE-Trans on the benchmark task of inter-patient brain MRI registration, while we believe that our NICE-Trans also could apply to other image registration applications (e.g., brain tumor registration [37]).

## 5      Conclusion

We have outlined a Non-Iterative Coarse-to-finE Transformer network (NICE-Trans) for medical image registration. Unlike the existing image registration methods, our NICE-Trans performs joint affine and deformable coarse-to-fine registration with a single network in a single iteration. The experimental results show that our NICE-Trans can outperform the state-of-the-art coarse-to-fine or transformer-based deep registration methods on both registration accuracy and runtime. Our study also suggests that transformers benefit registration in modeling inter-image spatial relevance while having limited benefits in learning intra-image representations.

**Acknowledgement.** This work was supported by Australian Research Council (ARC) under Grant DP200103748.

# Supplementary Materials for "Non-iterative Coarse-to-fine Transformer Networks for Joint Affine and Deformable Image Registration"

## A. Architecture Details

Table S1. Architecture details of the NICE-Trans used in the experiments.

| Encoder feature dimension | Decoder feature dimension | SwinTrans head number | SwinTrans window size |
|---|---|---|---|
| [8, 16, 32, 64, 128] | [256, 128, 64, 32, 16] | [16, 8, 4, 2, 0] | [5, 5, 5] |

## B. Registration Step Analysis

Table S2. Validation results of the NICE-Trans with different $L_a$ and $L_d$.

| $L_a$ | $L_d$ | DSC | NJD (%) | CPU runtime | GPU runtime |
|---|---|---|---|---|---|
| 0 | 3 | 0.590 | 0.038 | 4.32 | 0.468 |
| 1 | 3 | 0.606 | 0.030 | 4.54 | 0.482 |
| 2 | 3 | 0.607 | 0.032 | 4.70 | 0.488 |
| 0 | 4 | 0.593 | 0.034 | 4.52 | 0.480 |
| 1 | 4 | **0.610** | **0.025** | 4.69 | 0.486 |
| 2 | 4 | **0.610** | 0.028 | 4.75 | 0.490 |

**Bold**: the highest DSC and lowest NJD.

## C. Regularization Analysis

Table S3. Validation results of the NICE-Trans with different $\lambda$.

| NICE-Trans with $\lambda =$ | 0 | $10^{-5}$ | $10^{-4}$ | $10^{-3}$ |
|---|---|---|---|---|
| DSC | **0.625** | 0.614 | 0.610 | 0.598 |
| NJD | 2.17% | 0.102% | 0.025% | **0.003%** |

**Bold**: the highest DSC and lowest NJD.

## D. Qualitative Comparison

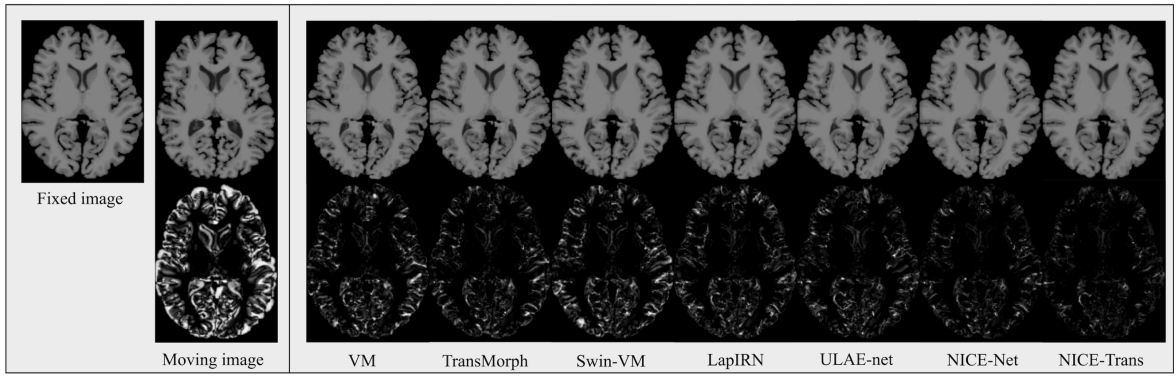

Fig. S1. Qualitative comparison between our NICE-Trans and comparison methods. Below each image is a difference map that shows the intensity differences between the corresponding image and the fixed image. A cleaner difference map indicates a better registration result.